\documentclass[a4paper]{llncs}
\usepackage{graphicx}
\usepackage{multirow}
\usepackage{booktabs}
\usepackage{subcaption}
\usepackage{graphicx}

\begin{document}
\title{JNLP Team: Deep Learning for Legal Processing in COLIEE 2020}
%
%

\author{Ha-Thanh Nguyen\inst{*}\inst{1} \and
Hai-Yen Thi Vuong\inst{*}\inst{1,2} \and
Phuong Minh Nguyen\inst{1} \and \\
Binh Tran Dang\inst{1} \and
Quan Minh Bui\inst{1} \and
Sinh Trong Vu\inst{1,4} \and
Chau Minh Nguyen\inst{1} \and
Vu Tran\inst{*}\inst{1} \and
Ken Satoh\inst{3} \and 
Minh Le Nguyen\inst{*}\inst{1} 
}
%
\authorrunning{\ }
%
\institute{Japan Advanced Institute of Science and Technology \and
VNU University of Engineering and Technology \and
National Institute of Informatics \and 
Banking Academy of Vietnam\\ 
\email{\{nguyenhathanh,phuongnm,
binhdang,quanbui,sinhvtr,chau.nguyen,vu.tran,nguyenml\}@jaist.ac.jp}, \email{yenvth@vnu.edu.vn}, \email{sinhvtr@bav.edu.vn}, \email{ksatoh@nii.ac.jp}}

\makeatletter{\renewcommand*{\@makefnmark}{}
\footnotetext{* Corresponding authors \\The works are conducted at Nguyen Lab and Interpretable AI Center - JAIST}\makeatother}

\maketitle              

\begin{abstract}
We propose deep learning based methods for automatic systems of legal retrieval and legal question-answering in COLIEE 2020. These systems are all characterized by being pre-trained on large amounts of data before being finetuned for the specified tasks. This approach helps to overcome the data scarcity and achieve good performance, thus can be useful for tackling related problems in information retrieval, and decision support in the legal domain. Besides, the approach can be explored to deal with other domain specific problems.

\end{abstract}
\section{\uppercase{Introduction}}
    COLIEE is an annual competition to find automated solutions in  the field of law. This competition is challenging because legal documents are often complex and require a high level of comprehension. The problems in law are even tricky for law experts. COLIEE tasks cover two of the most popular legal systems in the world, Case law and Civil law. COLIEE provides real data from the Canadian judicial system and the Japanese legal system.
    COLIEE organizes 4 tasks divided into 2 categories: retrieval and entailment. For retrieval tasks, the systems need to automatically find out the supporting cases of a given query case (Task 1) or the relevant articles of a given bar question (Task 3).
    For the entailment tasks, the systems need to find the paragraphs in a relevant case that entail a given decision (Task 2) or to conclude whether the statement of a given question is correct or incorrect (Task 4). 
   These tasks can be solved with various text processing methods. On one hand, over the years, systems using only lexical similarity of texts yield inferior performance. On the other hand, deep learning approaches start gaining superior performance recently.

    \subsubsection{Case Law} In COLIEE 2018, UBIRLED team use Tf-idf for ranking the candidate cases and abolish almost 75\% of lowest scoring candidate cases. Almost the same approach with UBIRLED team, UA team create a system which capture lexical matching between given base case and corresponding  candidate cases. This approach is applied for both Task 1 and Task 2 in COLIEE 2018 by several teams, but we can see that the performance is quite low with F1 of 0.3741 on Task 1 and 0.0940 on Task 2. JNLP team combines lexical matching and deep learning, which achieved state-of-the-art performance on Task 1 with F1 of 0.6545.
     
     In COLIEE 2019, several teams apply machine learning including deep learning to both of the tasks. JNLP team achieved the best system for Task 1 in COLIEE 2019 \cite{tran2019building} by using the approach similar to theirs in COLIEE 2018. In Task 2, their deep learning approach achieved lower performance compared to their lexical approach. However, team UA's combination of lexical similarity and BERT model achieved the best performance for Task 2 in COLIEE 2019~\cite{10.1145/3322640.3326741}.

    \subsubsection{Statute Law}
The retrieval task (Task 3) is often considered as a ranking problem with similarity features. In COLIEE 2019, JNLP\cite{jnlp_3_2019} chose Tf-idf as the main feature and calculated based on n-gram, noun phrase and verb phrase. DBSE\cite{dbse_3_2019} and IITP\cite{iitp_3_2019} based on BM25 to build their methods. DBSE used BM25 and Word2vec to calculate the similarity score. The KIS team\cite{kis_3_2019} represents article and query as a vector by generating a document embedding vector. Keywords are selected by Tf-idf and assigned high weight in the embedding process. 
    
    Regarding the entailment task (Task 4), approaches using deep learning are attracting more attention. In COLIEE 2019, KIS\cite{kis_4_2019} used predicate-argument structures to evaluate similarity. IITP\cite{iitp_3_2019} and TR\cite{tr_4_2019} apply BERT for this task. JNLP~\cite{jnlp_task4_coliee2019} classifies each query to follow binary classification based on big data.

\section{\uppercase{Case Law}}
    \subsection{Method}

    
    In legal technical terms, supporting cases in Task 1 are ``noticed" legal cases, which is considered to be relevant to a query case. Relevant cases are assumed to concern similar situations. A legal case document contains case details and metadata. The contents of the legal case are presented in a list of paragraphs that can be factual statements. 
    
    In Task 2, from given base case, an entailed fragment paragraph which extract from base case and the correctly paragraph from second case which support the entailed fragment. The mission is to discover which paragraphs within the second case support the entailed fragment from the base case.
    
    
    
    
    In the previous works, they usually tackle finding the supporting relationship between query-case/fragment and candidate-case/candidate-paragraph indirectly through similarity measures. Therefore, we want to build a model that could capture the supporting relationship directly. 
    
        \subsubsection{Supporting Model}
        We train the supporting model on a task of supporting text-pair recognition in case law.  The goal of the task is to determine whether a text supports a decision.  
        A small number of training data brings obstacles to the process of training deep neural models while collecting a large dataset for supporting relations can be challenging. Therefore, we designed some heuristics to automatically extract supporting text-pairs from case documents in Task 1's training data, and construct a ``weak-labeling" dataset. 
 The supporting model is based on the BERT base model, then finetuned on the supporting text-pair classification task. The BERT includes 768 hidden nodes, 12 layers, 12-attention heads, 110M parameters, 512 max input length.
        
        \subsubsection{Lexical Matching}
        The lexical similarity and the support relationship are distinct and can be complementary. Therefore, the lexical similarity is also potential in these tasks. We use BM25 to calculate lexical similarities. In Task 1, we separate the whole base cases and also candidate cases into set of paragraphs. We calculate the BM25 score where the query in this circumstance is each paragraph in set of paragraphs for given base case, and the documents is set of paragraphs of corresponding candidate cases. In Task 2, we have the same approach with Task 1 where the query is entailed fragment and the documents is set of candidate paragraphs. We use both of lexical score and supporting score as combine score:
        \begin{equation} \label{combine_score}
            score = \alpha * score_{supporting} + (1-\alpha) * score_{BM25}
        \end{equation}
        
        \subsubsection{Task 1}
        Deep learning models often consume much time and resources. 
        Therefore, we use lexical scoring to filter top $n$ cases from the given set of candidate cases and combine scores to identify supporting relations. All of the scores are calculated in paragraph level. Our supporting model does not use case-case supporting relation in the provided training data. 
        
        \subsubsection{Task 2}
        We try 3 settings as follows:
        \begin{enumerate}
            \item We directly use the already trained supporting model in Task 1 together with BM25 scoring for finding the support paragraphs for given entailed fragment.
            \item We enhance the system in the setting 1 by finetuning the supporting model on the training data of Task 2.
            \item We replace BM25 in the setting 2 with a BERT model finetuned on SigmaLaw dataset \footnote{https://osf.io/qvg8s/}. 
        \end{enumerate}
        
        \subsection{Task 1 Results}
        In COLIEE 2020, an short analysis is indicated in Table \ref{tab:task1_dataset_analysis}, every single case has more than 2.4K-token long in average, the number of paragraphs over 20. Not only about length but also the unbalance in the training and development set compare to the test set. We can see that the maximum number of words and paragraphs in test set are ten times more than training and development set. 
        \begin{table}
          \caption{Task 1 Data Analysis}
          \label{tab:task1_dataset_analysis}
          \begin{center}
            \begin{tabular}{|c|c|c|c|}
            \hline
            & train 2020 &dev 2020 &test 2020\\
            \hline
            Word/Doc & 2462 & 2443 & 3232\\ \hline
            Paragraphs/Doc & 23 & 23 & 28\\ \hline
            Maximum Word & 10827 & 10827 & 127263\\ \hline
            Maximum Paragraph & 119 & 119 & 1139 \\\hline
            Sample & 285 & 61 & 130 \\ \hline
            Candidate case & 57000 & 12200 & 26000 \\ \hline
            Average notice case & 5.21 & 5.41 & 4.89 \\
            \hline
            \end{tabular}
          \end{center}
        \end{table}

        As the results in the Table \ref{tab:task1_result}, we can see the lexical model (BM25) only achieves 0.5297 on F1. While the supporting model (W25 and W30) surpasses the lexical model by 0.1 on F1. Combining the lexical model and the supporting model (BMW25) get the best performance in our retrieval system.
        
        \begin{table}[!ht]
            \centering
            \caption{Results on Task 1.}  \label{tab:task1_result}
            \begin{tabular}{ |c| l | r | r | r | }
                \hline
                \textbf{Method} & \textbf{Hyper-parameters} &\textbf{Precision}  &\textbf{Recall}  &\textbf{F1} \\
                \hline 
                BM25& $\alpha = 0, topn = 25$  & 0.4071 & 0.7579 & 0.5297\\
                JNLP.W25 & $\alpha = 1, topn = 25$  & 0.5323 & 0.7893 & 0.6358\\
                JNLP.W30& $\alpha = 1, topn = 30$  & 0.5146 & 0.8050 & 0.6278\\
                JNLP.BM25& $\alpha = 0.85, topn = 25$  &  0.5603 & 0.7453 & 0.6397\\
                \hline
            \end{tabular}
        \end{table}

        Since Task 1 is a retrieval task, we want to optimize recall. That is the reason why our models obtain as many cases as possible and achieved high recall scores (the highest was 0.8). The average number of notice case in test set 2020 is less than that number in the train set and the dev set. This leads to the unbalance between recall and precision and low performance on F1. Although we didn't use the gold label from the training set, we achieved promising results in this retrieval task.

    \subsection{Task 2 Results}

        \begin{table}
          \caption{Task 2 Data Analysis}
          \label{tab:task2_data_analysis}
          \begin{center}
          \begin{tabular}{|c|c|c|c|}
            \hline
            & train 2020& dev 2020& test 2020\\\hline
            Case & 181 & 44 & 100\\\hline
            Candidate/Query & 32.12 & 32.91 & 36.72\\\hline
            Entailed fragment/Query & 1.12 & 1.02 & 1.25\\
          \hline
        \end{tabular}
          \end{center}
        \end{table}

        From Table \ref{tab:task2_result}, we notice that the lexical model achieve inferior performance, only 0.5764 F1. The setting 1 using the supporting model without Task 2 gold label performs better than the lexical model. Moreover, finetuning on the Task 2 gold label in the setting 2 further improves the performance and thus achieves the top performance of 0.6753 F1. Replacing BM25 by the BERT trained on SigmaLaw, however, reduces performance.  
        
        
        \begin{table}[!ht]
            \centering
            \caption{Result on Task 2.  \label{tab:task2_result}} 
            \begin{tabular}{ |c| l | r | r | r | }
                \hline
                \textbf{Method} & \textbf{Approaches} &\textbf{Precision}  &\textbf{Recall}  &\textbf{F1 score} \\
                \hline 
                BM25& Only BM25 score  & 0.6346 & 0.5280 & 0.5764\\
                JNLP.BMW & Setting 1  & 0.7000 & 0.5600 & 0.6222\\
                JNLP.BMWT& Setting 2  & \textbf{0.7358} & \textbf{0.6240} & \textbf{0.6753}\\
                JNLP.WT+L& Setting 3 & 0.6574 & 0.5680 & 0.6094\\
                \hline
            \end{tabular}
        \end{table}

\section{\uppercase{Statute Law}}
    \subsection{Task 3}
    \label{sec:task3_method}
 
    
    \begin{table}[!ht]
        \centering
        \small
        \caption{ Example of Task 3: pair of Query and Civil Code Article  \label{tab:example}} 
        \begin{tabular}{|l|l|p{0.72\textwidth}|}
            \hline 
            \multirow{8}{*}{Article} & Id  & ``303'' \\
                & Part & ``Part II Real Rights'' \\
                & Chapter & ``Chapter VIII Statutory Liens'' \\
                & Section & ``Section 1 General Provisions'' \\
                & Sumary line & ``(Content of Statutory Liens)'' \\
                & Content & ``The holder of a statutory lien has the rights to have that holder's own claim satisfied prior to other obligees out of the assets of the relevant obligor in accordance with the provisions of laws including this Act.'' \\
            \hline
            \multirow{2}{*}{Question} & Id  & ``H18-9-2'' \\
                &Content& ``Statutory real rights granted by way of security exist, but statutory usufructuary rights do not exist.'' \\
            \hline
        \end{tabular}
    \end{table} 
    
    As our observation, the challenge of this task is the long content of articles and the implication of meaning. 
    
\begin{figure*}[!ht]
        \caption{Sentence length distribution.} 
    
        \centering
        \begin{subfigure}[b]{0.49\textwidth}
            \includegraphics[width=\textwidth, keepaspectratio,  trim={1cm 0.7cm 1 1cm}, clip=true]{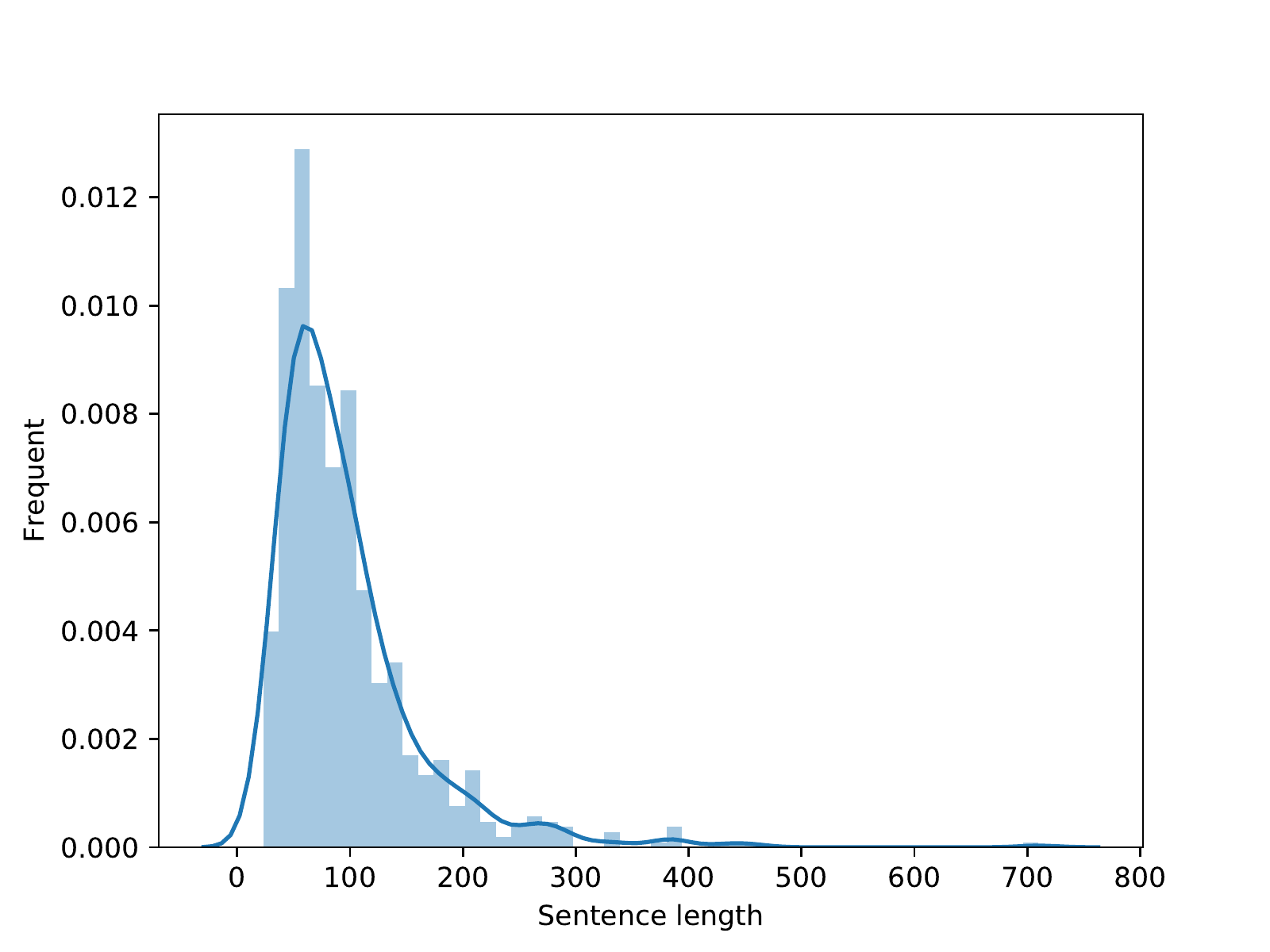}
            \caption{Articles}
            \label{fig:sentence_length_a}
        \end{subfigure}
        \begin{subfigure}[b]{0.49\textwidth}
            \includegraphics[width=\textwidth, keepaspectratio,  trim={1cm 0.7cm 1 1cm}, clip=true]{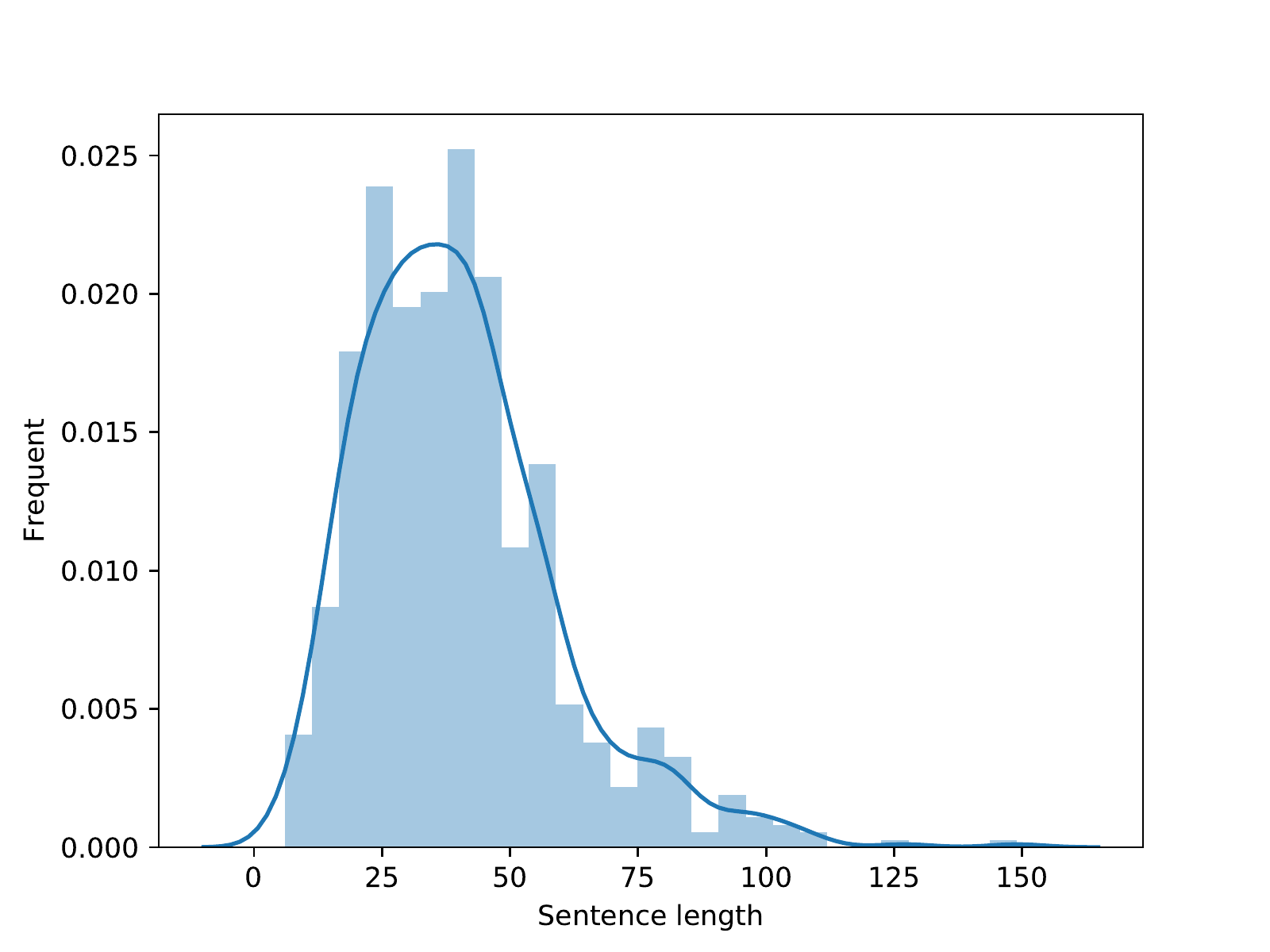}
            \caption{Questions}
            \label{fig:sentence_length_b}
        \end{subfigure}
    
        \label{fig:sentence_length}
    \end{figure*}

    The content of the articles in some cases is a paragraph that contains many sentences. 
    For example, the length distribution of articles and questions in the COLIEE Task 3 dataset is presented in Figure \ref{fig:sentence_length}. 
    The content of articles ($S_{i  \mid 1 \leq i \leq n} $) is the abstract description containing the conditions and relevant actions. The content of legal bar exam question ($Q$) is a particular case in daily life (Table~\ref{tab:example}). Therefore, these pairs $(Q; S_i)$ usually may not have much lexical overlap and it is difficult for methods using lexical matching and ranking to achieve high performance. 
    
    Recently, Transformer \cite{transformer} and two-stage (pre-training + fine-tuning) learning using BERT pre-trained model \cite{devlin2018bert} attracts a lot of attention because it is able to take advantage of pre-trained context before in many tasks in NLP. Therefore, in this work, we use BERT pre-trained model and fine-tune it with the pair $(Q; S_i)$ plays a role in two input sentences to classify the relevant between them. This approach changes the information retrieval problem into a classification problem that is similar to the MRPC subtask in the GLUE task \cite{devlin2018bert}. Our hypothesis is that the learned knowledge in BERT pre-trained model is helpful for meaning representation of both legal bar exam question ($Q$) and article ($S_i$) and improves the weak points of the approaches using word face representation.
     
    As we mentioned, we treat the task as classification to detect the relevant articles, so with each legal bar exam question, we need to classify each pair of the question with every article from the Civil Code. In fact, there are about 94\% of the legal bar exam questions have the number of relevant articles less than 3 while the total number of considering articles is more than 750. It is so unbalanced between relevant and non-relevant labels if we use all articles for pairing with each question. To prevent the unbalance problem, we used a pre-processing method that limits the number of non-relevant articles by selecting the $k$ (selected in fine-tune step) most appropriate articles using cosine similarity between Tf-idf vectors of the question and articles. In this way, we also reduce the unnecessary prediction for the classification model and diminish the bad impact of the unbalance problem.

    We build our system in two phases: data pre-processing and training/testing. 
    \begin{enumerate}
        \item \textbf{Data pre-processing}\quad  We filter the  $k$ best appropriate articles corresponding to the legal bar question. We build a Tf-idf vectorizer based on all articles in the Civil Code. After that, we encode legal bar questions and articles into vectors and use Cosine similarity to the rank of all articles. For each filtered article, we create a pair of it with the legal bar question and provided labels from COLIEE Task 3 dataset. Besides, to choose the $k$ value we do a small experiment to check the recall score of this step (Table \ref{tab:choose_k_tfidf}). Based on this result, we choose $k = 150$ for the training/testing phase.
        \begin{table}[!ht]
            \centering
            \small
            \caption{The dependency between $k$ and recall score.  \label{tab:choose_k_tfidf}} 
            \begin{tabular}{|l|r|r|r|r|r|r|r|}
                \hline 
                 $k$ & 30 & 50 & 70 & 100 & 120 & \textbf{150}  \\
                \hline
                Recall (\%) & 79.51 & 83.72 & 85.93 & 88.59 & 89.81 & \textbf{91.25}  \\
                \hline
            \end{tabular}
        \end{table} 
        \item \textbf{ Training/testing }\quad  In the training phase, we used BERT pre-trained model BERT-base-uncased) for fine-tuning data generated in the previous step. Besides, we assume that the BERT-base-uncased has strong knowledge in the general domain but weak in the legal domain. Therefore, we used all data from the Civil Code to fine-tune a special BERT model (BERT-CC) using the mask-language-model (MLM) \cite{devlin2018bert} and fine-tune the BERT-CC model for measuring relevance. Our hypothesis is that the BERT-CC and BERT-base-uncased learned different linguistic features, so the ensemble of these models can improve the recall score as well as performance. 
    \end{enumerate} 

        \subsubsection{Experiments} 
        
        \ 
        
    From the raw text file Civil Code, we used pattern matching to parse the structure of each article (example in Table \ref{tab:example}). To evaluate the performance of the model and optimize hyper-parameters, we separated a COLIEE Task 3 Questions dataset into two subsets train and dev. The dev subset contains about 10\% of the total that is all Questions having id field start with ``H29-*-*''. This subset plays a role as a development set to optimize hyper-parameters. To get the final submit results, we used the best setting found based on the dev subset and re-train on all original COLIEE Task 3 dataset. 

    Firstly, we conduct experiments to find the best setting of some hyper-parameters such as training epochs, max sequence length, etc. (Table \ref{tab:hyper-param-tunning}). 
    \begin{itemize}
        \item The runs 1, 2, 3 show that if we increase the number of training epochs, the model tends to increase the Precision score and decrease the Recall score. 
        \item The runs 3, 4, 5 show a similar variation, if we increase max sequence length, the model tends to increase the Precision score and decrease the Recall score. 
    \end{itemize} 
    Based on the F2 score, we found the best hyper-parameters setting is in ``Run 3''.  
    
    \begin{table}[!ht]
        \centering
        \caption{Optimize hyper-parameters of BERT-base-uncased on dev subset.  \label{tab:hyper-param-tunning}} 
        \begin{tabular}{ |c| l | r | r | r | }
            \hline
            \textbf{Run} & \textbf{Hyper-parameters} &\textbf{Prec}  &\textbf{Recall}  &\textbf{F2} \\
            \hline 
            1&epochs=10; max\_senquence\_len=230; lr=$2e^{-5}$ &  0.5857 & 0.5694 & 0.5726 \\
            2&epochs=5; max\_senquence\_len=230; lr=$2e^{-5}$ & 0.5211 & 0.5139 & 0.5153 \\
            3&epochs=3; max\_senquence\_len=230; lr=$2e^{-5}$ & 0.5500 & \textbf{0.6111} & \textbf{0.5978} \\
            4&epochs=3; max\_senquence\_len=300; lr=$2e^{-5}$ & 0.5270 & 0.5735 & 0.5636 \\
            5&epochs=3; max\_senquence\_len=400; lr=$2e^{-5}$ & \textbf{0.6000} & 0.4853 & 0.5046\\            6&epochs=3; max\_senquence\_len=230; lr=$1e^{-5}$ & 0.5417 & 0.5417 & 0.5417 \\
            \hline
        \end{tabular}
    \end{table} 
    
    Secondly, we conduct experiments to compare performance of different methods/models (Table \ref{tab:model_comparasion}). 
    \begin{itemize}
        \item \textbf{Tf-idf}.  This is the simple method using cosine to calculate the similarity of Question Tf-idf vector ($Q$) with each Article Tf-idf vector ($S_i$) and choose top two articles having the best scores.
        \item  \textbf{BERT-base-uncased (regression)}. In this run, we try to learn a regression model based on BERT-base-uncased. The results are the best performance after we optimized threshold values.   
        \item  \textbf{BERT-base-uncased}. This run used  BERT-base-uncased pretrained model to fine-tune.  
        \item  \textbf{BERT-CC}. As mentioned before, this model we fine-tune a language model with MLM task from BERT-base-uncased and continue to fine-tune to detect relevant articles.
        \item  \textbf{BERT-ensemble}. In this model, we ensemble the result of 2 models:  BERT-base-cased and BERT-CC, the article is relevant to the legal bar question if one of these models predicts True.
    \end{itemize}
    
    These results show that the BERT-CC learned more information in the legal domain as our expectation with the highest precision. Besides, the BERT-ensemble can utilize the result of BERT-CC and BERT-base-uncased to improve the performance (recall and F2) of the overall system.
    
    \begin{table}[!ht]
        \centering
        \small
        \caption{The comparasion performance of models on dev subset.  \label{tab:model_comparasion}} 
        \begin{tabular}{| l | r | r | r | }
            \hline
            \textbf{Model} &\textbf{Prec}  &\textbf{Recall}  &\textbf{F2} \\
            \hline 
            Tf-idf (Top 2 best articles) &  0.3276 & 0.5278 &  0.4703 \\
            \hline 
            BERT-base-uncased (regression)& 0.4819 & 0.5882&0.5634 \\
            \hline 
            BERT-base-uncased & 0.5500 & 0.6111 & 0.5978 \\
            BERT-CC & \textbf{0.5694} & 0.6029 & 0.5959 \\
            BERT-ensemble & 0.5111 & \textbf{0.6389} & \textbf{0.6085} \\
            \hline
        \end{tabular}
    \end{table} 
    \subsubsection{Performance on the test data}  The performance on test set  is 2nd in the COLIEE-2020 competition, showed on Table \ref{tab:task3_result}.  Similar to the result on the dev subset, the model BERT-ensemble achieves the best result in comparing with BERT-CC and BERT-base-uncased. 
    \begin{table}[th]
    \centering
      \caption{Task 3 - final result on test set}
      \label{tab:task3_result}
      \begin{tabular}{|l|c|c|c|c|c|c|c|}
        \hline
        \textbf{Run}& \textbf{Prec}	&\textbf{Recall}& \textbf{F2}&		\textbf{MAP}	&\textbf{R5}	&\textbf{R10}	&\textbf{R30}\\
        \hline
        JNLP BERT-ensemble &\textbf{0.5766} &\textbf{0.5670}&\textbf{	0.5532 }  &0.6618 &0.6857 &0.7143 &0.7786\\\hline
        JNLP BERT-CC	&0.5387 &0.5268 & 0.5183	&0.6847 &0.7143	&0.7714 &0.7929\\\hline
        JNLP BERT-base-uncased  &0.5409	&0.5268 &      0.5149		&0.6847	&0.7143	&0.7714	&0.7929\\
      \hline
    \end{tabular}
    \end{table}
    We show the improving predictions of  BERT-CC when comparing with BERT-base versions in Table \ref{tab:compareCCbase}. When comparing with BERT-CC, BERT ensemble version predicts more than 20 relevant pairs of Question and Article but there are only 7 correct relevant pairs learned by BERT-base. When compare with BERT-base, BERT ensemble version predicts more than 7 relevant pairs  and there are 5 correct relevant pairs that learned by BERT-CC.  We hypothesis that the BERT-CC model is different from BERT-base to pay attention to the different important words in a sentence with higher accuracy.        
    
    \begin{table}
      \caption{Improving predictions of BERT-CC missed by BERT-base. \label{tab:compareCCbase}}
      \begin{center}
          
      \begin{tabular}{|l | p {10cm} |}
        \hline
        \textbf{Pair}& \textbf{Content} \\\hline
        Q: \textit{R01-1-E} & An \ldots such consent or the permission of family court in lieu thereof.\\ 
        A: \textit{17} & Part I General Provisions  \ldots  Requiring Person to Obtain Consent of Assistant) \ldots  obtain the consent of the person's assistant \ldots \\ 
       \hline
        Q: \textit{R01-3-O}  & When An unauthorized agent acted believing that \ldots  \\ 
        A: \textit{117} &  \ldots Section 3 Agency   (Liability of Unauthorized Agency)  (1) A person who concludes \ldots \\ \hline
        Q: \textit{R01-10-O}  & A, B and C co-own Building X at the rate of one-third each. \ldots  \\ 
        A: \textit{254} &  \ldots Ownership Section 3 Co-Ownership   (Claims on Property in Co-Ownership)  A claim that one of the co-owners \ldots \\ \hline
        Q: \textit{R1-20-I}  & In cases where the seller of the specified things retains them at the place \ldots  the tender of the performance shall be sufficient \ldots \\ 
        A: \textit{484} &  \ldots Unless a particular intention is manifested with respect to the place where the performance should take place \ldots \\
       \hline
    \end{tabular}
    
      \end{center}
    \end{table}


    \subsection{Task 4}
    \label{sec:task4_method}
    The goal of Task 4 is to find answers to given bar questions. Bar questions are the questions to judge paralegal examinees, making sure they are qualified to practice legal activities. The reasoning sequence of an examinee when perceiving a question is as follows: The examinee reads and understands the question, delineates the scope of relevant legal knowledge, finds the articles that reject or support the proposition in question and decide the final answer. There are two possible outcomes, Yes and No, which the system has to decide on for each question.
    
    A system designer's natural thought is to use the result of Task 3 as input to Task 4 in the form of a Texture Entailment problem. However, this approach may cause the outcome of Task 4 to be biased into the selection of relevant articles in Task 3.
    
    Also, thinking about the paralegal's process of finding answers in the bar exam, we suppose that the examinee's thinking process might be different from conventional system design. A good examinee would not try to remember every word in the civil code to conduct inference. They will try to abstract out the principles in the law as well as situations they have learned earlier.
    
    From the above observation, in terms of system design, we use two approaches: using the results of Task 3 together with the questions as input to the texture entailment problem and using only the content of questions as inputs of the classification problem. With the second approach, our hypothesis is that deep learning systems can automatically generate latent patterns at an abstract level to answer questions.
    
    In the first approach, we use BERT-base-uncased to implement texture entailment. The question was linked to the relevant article by a \textit{[SEP]} token and fed into BERT model. On our own development set, we found that using gold data with extra relevant articles extracted by Tf-idf directly from Civil Code brings better performance than using only gold data from Task 3. 
    For each question, we applied the textual entailment with a new data named \textit{Tf-idf extra data} that contains gold data and 2 top articles returned by \textit{Tf-idf}. The system decides the answer is Yes if at least one pair is classified as positive textual entailment.
    
    The second approach is firstly proposed by Nguyen et al.~\cite{jnlp_task4_coliee2019} in COLIEE-2019. By converting the texture entailment problem into the lawfulness classification problem, the data used to build the model becomes more abundant.
    In this approach, the model is trained directly from law sentences, bar questions, and their variants with two labels: Yes and No. Data sources are the Civil Code and previous years' bar questions provided by the COLIEE organizers.
    
    We use two variants of BERT in this approach, pretrained BERT-base-uncased by Google and BERT Law trained from scratch by a corpus containing 8.2M sentences (182M words) from American case data in English.
    
    Both variations contain 768 hidden nodes, 12 layers, 12-attention heads, 110M parameters. They are then finetuned and validate on the dataset, including their law sentences, bar questions, and their augmented samples. The maximum length in our setting is 128.
    
    \begin{figure}[h]
      \centering
      \includegraphics[width=.6\linewidth]{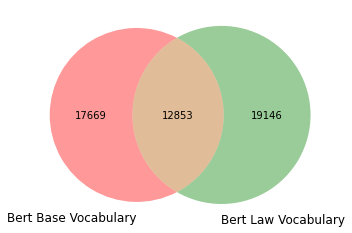}
      \caption{The cardinality of two vocabulary sets of Bert Base and Bert Law.}
      \label{fig:venn_diagram}
    \end{figure}

    These two vocabulary sets share only 12,853 tokens. 17,669 tokens appear only in BERT Base vocabulary and 19,146 tokens appear only in BERT Law vocabulary. This is an interesting figure showing the difference in how the two models of the same architecture learn to use vocabulary in the general domain (BERT Base) and the specialized domain (BERT Law).


\subsubsection{Textual Entailment Approach} 
In the textual entailment approach, we conduct an experiment to choose the more suitable source of relevant articles: original gold data of Task 3 or Tf-idf extra data. The dataset to train and validate the model is the question bars from the previous years provided by the organizers. The development set is 10\% of the entire dataset. Table \ref{tab:article_source_task4} presents the performance on the two sources.

\begin{table}
  \caption{Using relevant articles retrieved by \textit{Tf-idf} yeilds better performance}
  \label{tab:article_source_task4}
  \begin{center}
      
  \begin{tabular}{|l|c|}
    \hline
    \textbf{Data Source}& \textbf{Accuracy}\\
    \hline
    Gold data of Task 3           &0.5278\\\hline
    Tf-idf extra data         &\textbf{0.5417}\\
  \hline
\end{tabular}

  \end{center}
\end{table}

Using the \textit{Tf-idf extra data} yields 1.39 \% higher than the result from the gold data of Task 3. This result interestingly suggests the possibility that there exists a set of relevant articles other than gold data of Task 3 that can aid in deciding answers to bar questions. This is consistent with the actual phenomenon that occurs in the litigation process, lawyers can come up with different arguments to defend or reject the same opinion.

\subsubsection{Lawfulness Classification Approach} 
Compared with the Textual Entailment approach, the Lawfulness Classification approach has much more abundant data. All augmented data from law sentences and previous years' bar questions are 5000 samples. We also use 10\% of the dataset as the development set. Table \ref{tab:bert_compare_task4} shows the performance of BERT Law and BERT-base-uncased in the development set.

\begin{table}
  \caption{BERT Law overperforms BERT Base}
  \label{tab:bert_compare_task4}
  \begin{center}
      
  \begin{tabular}{|l|c|}
    \hline
    \textbf{Model}& \textbf{Accuracy}\\
    \hline
    BERT Base          &0.7784 \\\hline
    BERT Law         &\textbf{0.8168}\\
  \hline
\end{tabular}

  \end{center}
\end{table}

Table \ref{tab:article_source_task4} and Table \ref{tab:bert_compare_task4} report results evaluated on two different development sets of two different problems. The conclusions we draw from the above two tables are that: on our development set, using data from retrieved from \textit{Tf-idf} will be better in the Textual Entailment problem and using BERT Law will be better in Lawfulness Classification problem. These are important information for us to choose the candidates for our final runs.

\subsubsection{Performance on the gold data} 
Our results on gold data of Task 4 are shown in the Table \ref{tab:task4_result}. This table of results shows that the second approach outperforms the first. In addition, BERT Law creates a significant gap compared to Google's pretrained BERT-base-uncased. The model comparison result is consistent with our experimental results on our development set. 

\begin{table}
  \caption{Final runs result on gold data of Task 4}
  \label{tab:task4_result}
  \begin{center}
      
  \begin{tabular}{|l|c|c|}
    \hline
    \textbf{Run}& \textbf{Correct}& \textbf{Accuracy}\\
    \hline
    JNLP BERT Law           &81 &\textbf{0.7232}\\\hline
    JNLP BERT Base          &63	&0.5625\\\hline
    JNLP Tf-idf BERT	Base    &62	&0.5536\\
  \hline
\end{tabular}

  \end{center}
\end{table}
\subsubsection{Discussion} 
According to our observations, the test data for this task in previous years are usually small, leading to significant variability of the same model from the development set to the test set and from year to year.
Therefore, choosing the best model based on a development set does not guarantee it will work well on test data. Despite this, the COLIEE Task 4 results this year brought us by surprise with the fact that BERT Law made a big gap compared to other models.

As our observation from the experimental results, pretrained deep learning models can perform well on comprehensive tasks like law. And it achieves good results with the right training methods as well as good data in both quantity and quality.




\section{\uppercase{conclusion}}
In this paper, we reported our methods and results in using different techniques with Deep learning for the COLIEE 2020 comprehensive legal text processing tasks. Our approaches are based on trained deep learning models with large amounts of data to support decisions for retrieval and entailment problems. Experimental results show that our systems yield competitive performance. 

The results demonstrate that it is feasible to use the deep learning models for advanced tasks such as law document processing as long as the appropriate approach is taken.
In future works, we will investigate to build different models with high stability, applicable to real-life applications.

\section*{Acknowledgments}
This work was supported by JST CREST Grant Number JPMJCR1513 and JSPS KAKENHI Grant Number JP17H06103. Vuong and Chau
were supported in part by the Asian Office of Aerospace R\&D (AOARD), Air Force Office of Scientific Research (Grant no. FA2386-19-1-4041)

\bibliographystyle{splncs04}
\bibliography{ref}







\end{document}